%% file: main.tex
\documentclass[runningheads]{llncs}
\usepackage{graphicx}

\usepackage{amsmath,amssymb} 
\usepackage{color}
\usepackage{times}
\usepackage{epsfig}
\usepackage{subfig}
\usepackage{verbatim}
\usepackage{blindtext}
\newcommand{\R}{\mathbb{R}}
\def\eg{{\em e.g.}\@\xspace}
\def\etal{{\em et al.}\@\xspace}
\def\etc{{\em etc.}\@\xspace}
\def\ie{{\em i.e.}\@\xspace}

\graphicspath{{./figs/}}

\usepackage[table]{xcolor}
\usepackage[outline]{contour}

\begin{document}

\title{Semantic bottleneck for computer vision tasks}
\titlerunning{Semantic bottleneck} 


\author{Maxime Bucher\inst{1,2} \and
St\'ephane Herbin\inst{1} \and
Fr\'ed\'eric Jurie\inst{2}}
%

\authorrunning{M. Bucher et al.} 


\institute{ONERA, Universit\'{e} Paris-Saclay, FR-91123 Palaiseau, France \and
Normandie Univ, UNICAEN, ENSICAEN, CNRS }

\maketitle

\input{2-abstract.tex}
\input{3-introduction.tex}
\input{4-related_work.tex}

\input{5-method.tex}

\input{6-experiments.tex}
\input{7-conclusions.tex}
\clearpage
\bibliographystyle{splncs}
\bibliography{0-bibfile.bib}
\clearpage
\end{document}

%% file: 2-abstract.tex
\begin{abstract}
This paper introduces a novel method for the representation of images that is semantic by nature,  addressing the question of computation intelligibility in computer vision tasks. More specifically, our proposition is to introduce what we call a \emph{semantic bottleneck} in the processing pipeline, which is a crossing point in which the representation of the image is entirely expressed with natural language, while retaining the efficiency of numerical representations. We show that our approach is able to generate semantic representations that give state-of-the-art results on semantic content-based image retrieval and also perform very well on image classification tasks. Intelligibility is evaluated through user centered experiments for failure detection.
\end{abstract}

%% file: 3-introduction.tex
\section{Introduction}

\begin{figure}[tb]
\centering
\includegraphics[scale=0.30]{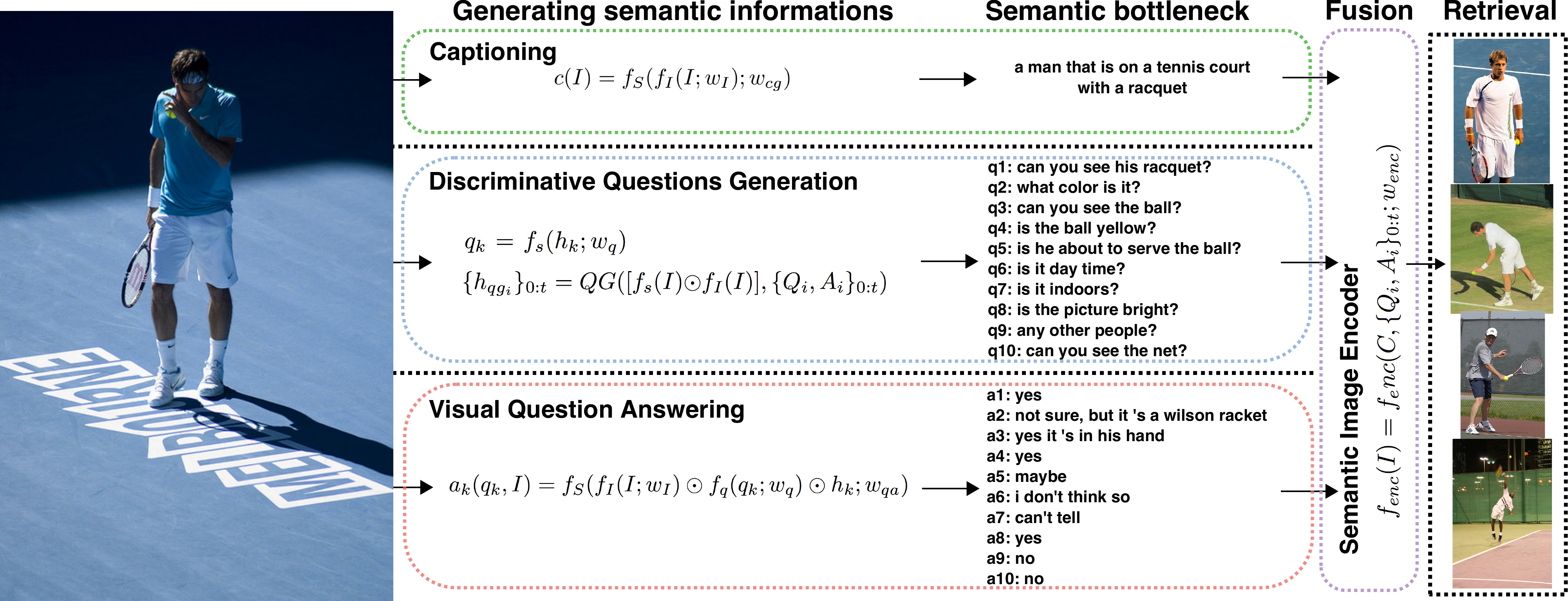}
\caption{Semantic bottleneck approach: images are replaced by purely but rich textual representations, for tasks such as multi-label classification or image retrieval.}
\label{fig:visu_gen}
\end{figure}

Image-to-text tasks have made tremendous progress since the advent of deep learning approaches (see \eg, \cite{das2016visual}). The work presented in this paper builds on these new types of image-to-text functions to evaluate the capacity of textual representations to semantically and fully encode the visual content of images for demanding applications, in order to allow the prediction function to host a \emph{semantic bottleneck} somewhere in its processing pipeline (Fig.~\ref{fig:visu_gen}). 
The main objective of a semantic bottleneck is to play the role of an \emph{explanation}  of the prediction process since it offers the opportunity to examine meaningfully on what ground will further predictions be made, and potentially decide to reject them either using human common sense knowledge and experience, or automatically through dedicated algorithms. Such an explainable semantic bottleneck instantiates a good tradeoff between prediction accuracy and interpretability \cite{gilpin2018explaining}.

Reliably evaluating the quality of an explanation is not straightforward \cite{doshi2017roadmap,biran2017explanation,gilpin2018explaining,ras2018explanation}. 
In this work, we propose to evaluate the explainability power of the semantic bottleneck by measuring its capacity to detect failure of the prediction function, either through an automated detector as \cite{zhang2014predicting}, or through human judgment. 
Our proposal to generate the surrogate semantic representation is to associate a global and generic textual image description (caption) and a visual quiz in the form of a small list of questions and answers that are expected to refine contextually the generic caption. The production of this representation is adapted to the vision task and learned from annotated data.




The main contributions of this paper are:
{\em (i)} The design of two processing chains for content-based image retrieval and multi-label classification hosting a semantic bottleneck;
{\em (ii)} An original scheme to select sequentially a list of questions and answers to form a semantic visual quiz;
{\em (iii)}  A global fusion approach jointly exploiting the various components of the semantic representation for image retrieval or multi-label classification;
{\em (iv)}  A complete evaluation on the MS-COCO database exploiting Visual Dialog annotations \cite{das2016visual} showing that it is possible to enforce a semantic bottleneck with only 5\% of performance loss on multi-label classification, but a 10\% performance gain for image retrieval, when compared to image feature-based approaches;
{\em (v)} An evaluation of the semantic bottleneck explanation capacity as a way to detect failure in the prediction process and improve its accuracy by rejection.

%% file: 4-related_work.tex
\section{Related works}
\label{sec:related_work}


\emph{Extracting semantic information from images.} The representation of images with semantic attributes has received a lot of attention in the recent literature. 
However, with the exception of the DAP model \cite{Lampert:2014fs}, which is not performing very well, such  models produce vector representations that are not intelligible at all.  In contrast, image captioning \cite{Vinyals:2016jd,Lu:2017km} is by nature producing intelligible representations and can be used to index images. As an illustration, Gordo \etal \cite{Gordo_2017_CVPR} addressed the task of retrieving images that share the same semantics as the query image using captions. Despite the success of such recent methods, it has been observed \cite{Dai:2017ww} that such approaches produce captions that are similar when they contain one common object, despite their differences in other aspects. Addressing this issue, \cite{Dai:2017wn} proposed a  contrastive learning method for image captioning encouraging distinctiveness, while maintaining the overall quality of the generated captions. Another way to enrich the caption is to generate a set of questions/answers such as proposed in the Visual Dialog framework \cite{das2016visual,Seo:2017ve,Lu:2017tb}. This is what we propose to do by learning how to build dialogs complementary to image captions. 

 
\emph{Transferring information from other domains.} Producing semantic description of images in natural languages is barely possible without transferring semantic information --  expressed as semantic attributes, natural language expressions, dictionaries, \etc -- from auxiliary datasets containing such information to novel images. This is exactly what Visual Question Answering models can do, the VQA challenge offering important resources, in the form of semantic images, questions, or possible answers. Research on VQA has been very active during the last two years. \cite{malinowski2015ask} proposed an approach relying on  Recurrent Neural Network using Long Short Term Memory (LSTM). In their approach, both the image (CNN features) and the question are fed into the LSTM to answer a question about an image. Once the question is encoded, the answers can be generated by the LSTM. \cite{lin2016leveraging} study the problem of using VQA knowledge to improve image-caption ranking. \cite{zhu2016visual7w}, motivated by the goal of developing a model based on grounded regions, introduces a novel dataset that extends previous approaches and proposes an attention-based model to perform this task. On their side, \cite{jabri2016revisiting}  proposed a model receiving the answers as input and predicts whether or not an image-question-answer triplet is correct. Finally, \cite{wu2016ask} proposes another VQA method combining an internal representation of the image content  with the information extracted from general knowledge bases, trying to make the answering of more complex questions possible.

\emph{Producing intelligible representations.} The ubiquitousness of deep neural networks in modern processing chains, their structural complexity and their opacity have motivated the need of bringing some kind of intelligibility in the prediction process to better understand and control its behavior.
The vocabulary and concepts connected to intelligibility issues are not clearly settled. (explanation, justification, transparency, \etc) 
Several recent papers have tried to clarify those expressions \cite{lipton2016mythos,doshi2017roadmap,doran2017does,biran2017explanation,hohman2018visual,gilpin2018explaining,guidotti2018survey,ras2018explanation} and separate the various approaches in two goals: build interpretable models and/or provide justification of the prediction.
\cite{zhang2018interpreting} for instance, described an interpretable proxy (a decision tree) able to explain the logic of each prediction of a pretrained convolutional neural networks. The generation of explanations as an auxiliary justification has been addressed in the form of a visual representation of informative features in the input space, usually heat maps or saliency maps \cite{selvaraju2017grad,rajani2017using,montavon2018methods}, as textual descriptions \cite{hendricks2016generating}, or both \cite{park2018multimodal}. A large body of studies \cite{mahendran2015understanding,zeiler2014visualizing,dosovitskiy2016inverting,olah2018building} have been interested in  visually revealing the role of deep network layers or units.
Our semantic bottleneck approach fuses those two trends: it provides a directly interpretable representation, which can be used as a justification of the prediction, and it forces the prediction process itself to be interpretable in some way, since it causally relies on an intermediate semantic representation. 




\emph{Evaluating explanations.} The question of clearly evaluating the quality or usability of explanations remains an active problem.
\cite{selvaraju2017grad} described a human-centered experimental evaluation assessing the predictive capacity of the visual explanation. \cite{montavon2018methods} proposed to quantify explanation quality by measuring two desirable features: continuity and selectivity of the input dimensions involved in the explanation representation. \cite{zhang2017interpreting} and \cite{bau2017network} described geometric metrics to assess the quality of the visual explanation with respect to landmarks or objects in the image. \cite{kindermans2017reliability} questioned the stability of saliency based visual explanations by showing that a simple constant shift may lead to uninterpretable representations.
In our work, we take a dual approach: rather than evaluating the capacity of the explanation to be used as a surrogate or a justification of an ideal predictive process, we evaluate its quality as an ability of detecting bad behavior, \ie detect potential wrong predictions.



\emph{Generating distinctive questions.} If the generation of questions about text corpora has been extensively studied (see \eg, \cite{serban2016generating}), 
the generation of questions about images has driven less attention. We can, however, mention the interesting work of \cite{li2017learning}  where discriminative questions are produced to disambiguate pairs of images, or \cite{mostafazadeh2016generating} which  introduced the novel task (and dataset) of visual question generation.  We can also mention the recent work of Das \etal \cite{DasKMLB17} which bears similarity with our approach but differs in the separation between the question generator and semantic representation encoder, and is not applied to the same tasks. Our work builds on the observation made by \cite{ganju2017s,zhuknowledge} -- questions that are asked about an image provide information regarding the image and can help to acquire relevant information about an image -- and proposes to use automatically generated discriminative questions as cues for representing images.








%% file: 5-method.tex
\section{Approach}
\label{sec:approach}
\begin{figure}[tb]
    \centering
    \subfloat[VQA oracle model]{{\includegraphics[width=0.45\textwidth]{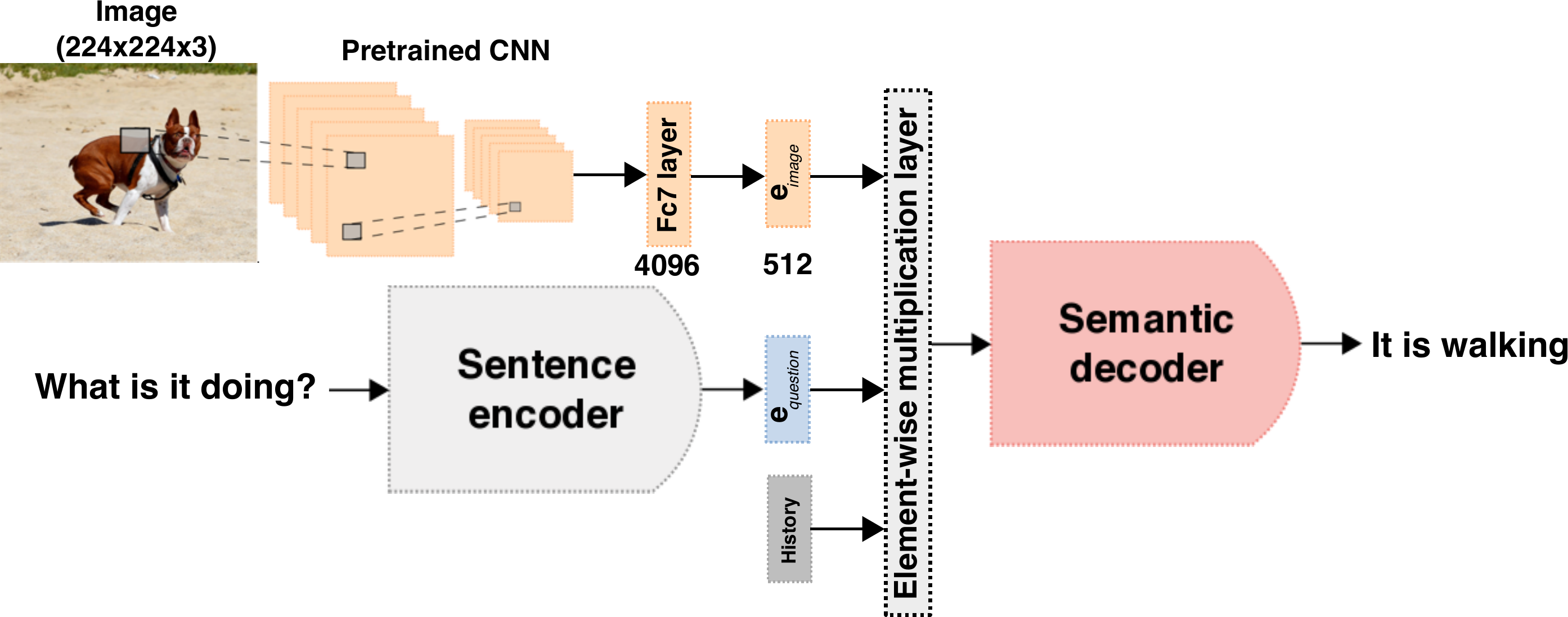} }}%
    \qquad
    \subfloat[Captioning oracle model]{{\includegraphics[width=0.44\textwidth]{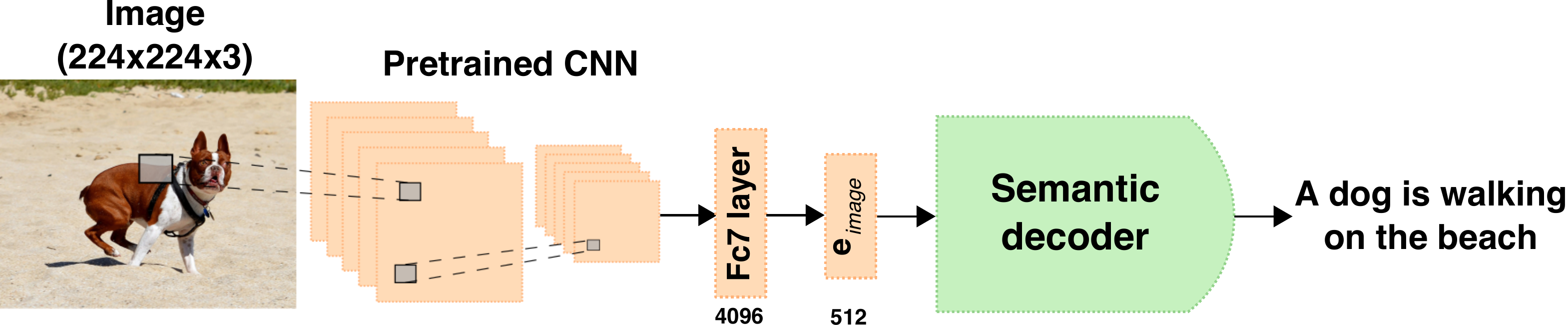} }}%
      \qquad
    \subfloat[Discriminative Question Generator]{{\includegraphics[width=0.45\textwidth]{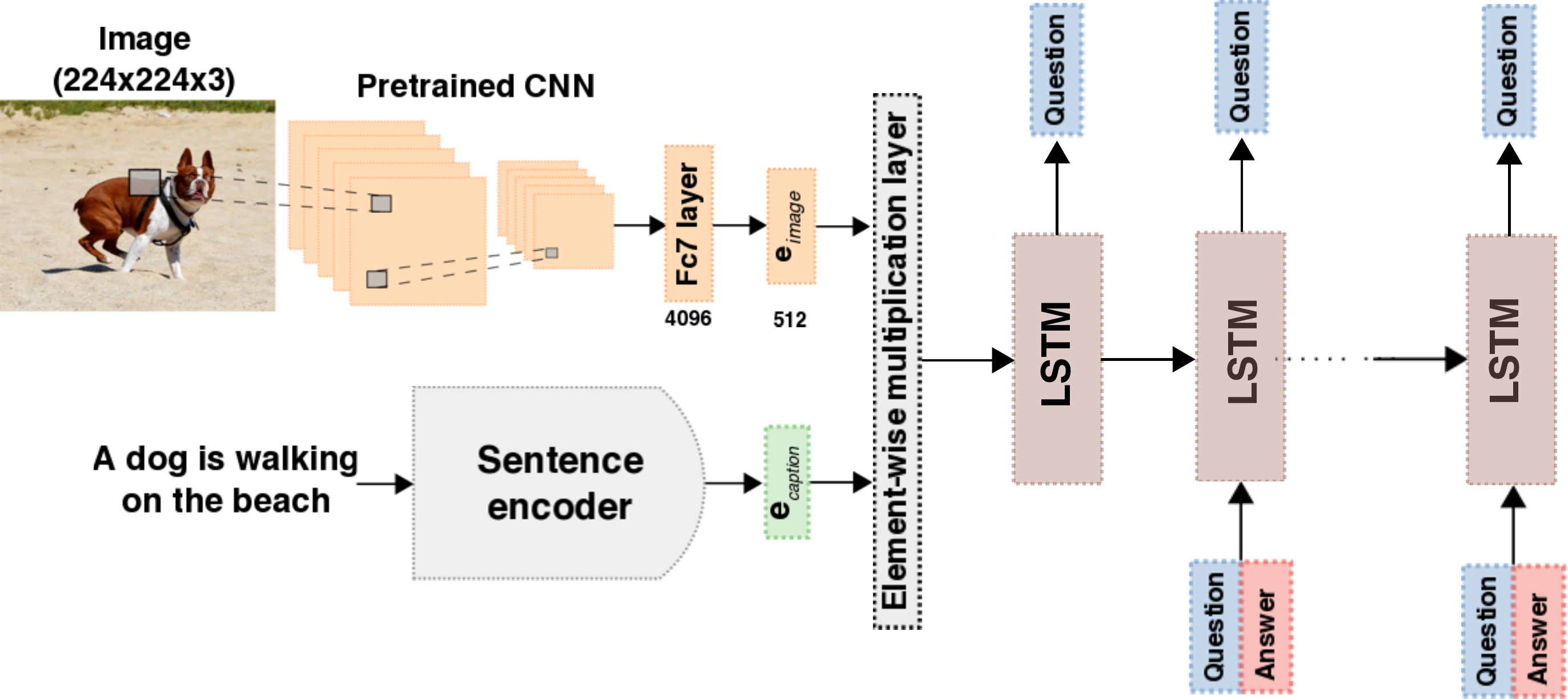} }}%
    \qquad
    \subfloat[Semantic Representation Encoder]{{\includegraphics[width=0.45\textwidth]{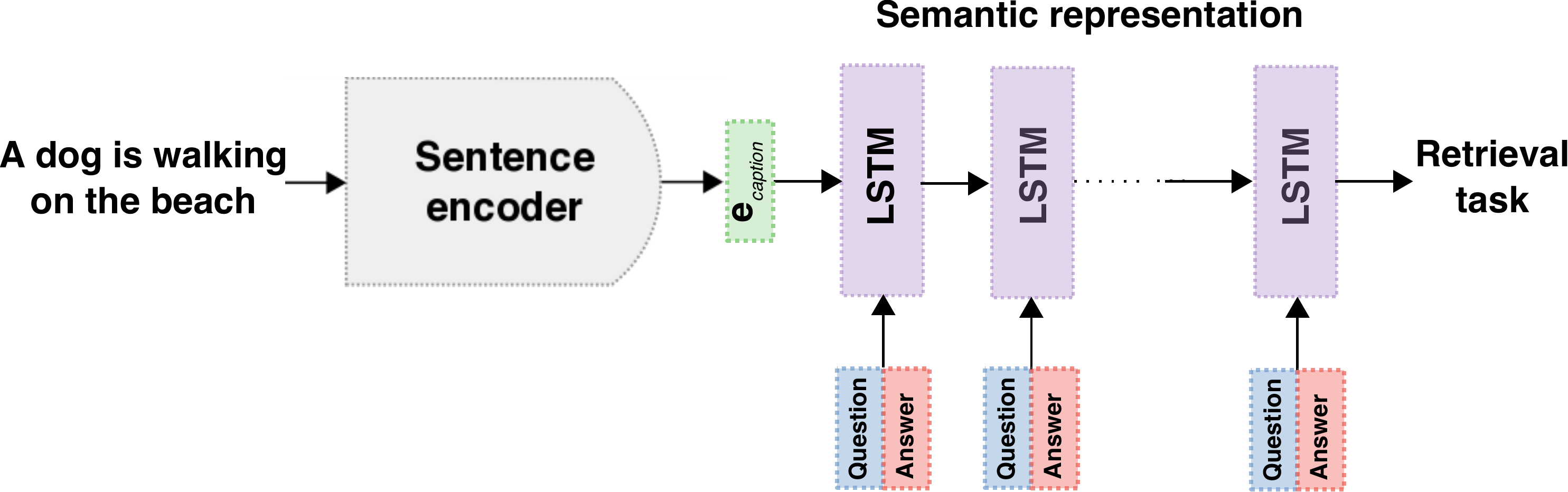} }}%
    \qquad
      \caption{Functional diagrams of the various components of the global algorithm. }%
    \label{fig:blocks}%
\end{figure}

This paper proposes a method allowing to turn raw images into rich semantic representations --- a semantic bottleneck --- which can be used alone (without using the image itself) to compare images or classify them. At the heart of this image encoder is a process for generating an ordered set of questions related to the image content, which are used, jointly with the answers to these questions, as an image substitute. Questions are generated sequentially, each question (as well as its answer) inducing the next questions to be asked. Such a set of questions/answers  should semantically represent the visual information of images and be useful for disambiguating image representations in retrieval and classification task.  Answering the question is done with a VQA model, used as an oracle, playing the role of a fine-grained information extractor.  
  Furthermore, this sequence of Q/A is designed to be complementary to an image caption which is also automatically generated. We can use the analogy of human reasoning, starting with the image caption as a starting point and asking question to an oracle to get iteratively more information on the image. The proposed visual dialog allows to enrich image caption representations and leads to a stronger semantic representation. 
Finally, captions and visual dialogs are combined and turned into a compact representation that can be used easily to compare images (for retrieval task) or to infer class labels (classification tasks). 

Consequently, our model is composed of two main components: i) a discriminative visual question generator ii) an encoding block taking  Q/A and captions as inputs and producing an image representation. These two blocks, trained end-to-end, rely on two oracles: i) an image caption generator which visually describes images with natural language sentences. ii) a visual question answering model capable of  answering free-from questions related to images. We call these  last 2 parts of the model {\em oracles} as they are trained independently of the main tasks and used as external knowledge base.

\subsection{Vector space embedding, encoders, decoders}
\label{sec:semantic_encoder_decoder}
The core objective of our approach is to generate semantic expressions in natural language (questions, answers or captions) that could represent images compactly and informatively. 
We define the various natural language elements as sequences of words from a fixed vocabulary of words and punctuation $\{w_0,\ldots,w_{n_w}\}$, where $w_{0}$ has the special meaning of marking the end of all phrases as a ``full stop''. The space of all possible sequences in natural language is denoted by $\cal P$. Any caption ($c$), question ($q$) or answer ($a$) belongs to the same set $\cal P$. 

Most of learning based algorithms exploit vector space representations as inner states, or in the optimized criterion. A first issue is therefore to make possible the embedding of natural language expressions, and also images, into a vector space. For simplicity of design, we made all the necessary embeddings belong to a $S$-dimensional real valued vector space. Typically, to give an order of magnitude, we took $S=512$ in our experiments. We therefore define semantic {\em encoders} as mappings from $\cal P$ to $\R^S$, and semantic {\em decoders} or {\em generators} as mappings from the vector space $\R^S$ to $\cal P$.



\paragraph{Image encoder}
The first element to be encoded is the source image $I \in \cal{I}$. The encoder (denoted as $f_I$) is a mapping $f_I: \cal{I}\rightarrow \R^S$, provided by last FC layer (fc7) of a VGG-VeryDeep-19 network~\cite{simonyan2014very} (pre-trained on Imagenet \cite{russakovsky2015imagenet}) followed by a non-linear projection (fully connected layer + tanh non-linearity unit)  reducing the dimensionality of the fc7 features (4096-d) to $S$. The parameters of the non-linear projection, denoted as $w_I$, are the only parameters of this embedding that have to be learned, the parameters of the VGG-VeryDeep-19 being considered as fixed. We write $f_I(I;w_I)$ to make apparent the dependency on the parameters $w_I$, when needed.

\paragraph{Natural language encoder}
This encoder maps any set of words and punctuation $s \in \cal P$ (captions, questions, answers) to the embedding space $\R^S$. We will use 3 different natural language encoders in the global algorithm, for captions, questions and answers, all sharing the same structure and the same weights. We hence refer to them using the same notation ($f_p$). The encoder uses a standard Long Short Term Memory network (LSTM) \cite{hochreiter1997long}.
Used as an encoder, the LSTM is simply a first order dynamic model:
$y_t=\mathrm{LSTM}_p(y_{t-1}, p_t)$ 
where $y_t$ is the concatenation of the long-term cell state and the short-term memory, and $p_t$ is the current input at time $t$. 
Given a natural language sequence of words $p = \{p_t\}_{t=1:T_p}$, its encoding $f_p(p)$ is equal to the memory state $y_t$ after iterating on the LSTM 
$T_p$ times and receiving the word $p_t$ at each iteration. Denoting $w_p$ the set of weights of the LSTM model, the natural language embedding is therefore defined as:
$y_{T_p} = f_p(p;w_p)$
with the initial memory cell $y_0 = 0$.

In practice, instead of using words by their index in a large dictionary, we encode the words in a compact vector space, as in the word2vec framework \cite{mikolov2013efficient}. We found it better since synonyms can have similar encodings. More precisely, this local encoding is realized as a linear mapping $w_{w2vec}$, where $w_{w2vec}$ is a matrix of size $ n_{w2vec} \times n_w$ and each original word is encoded as a one-hot vector of size $n_w$, the size of the vocabulary. The size of the word embedding $n_{w2vec}$ was 200 in our experiments. This local word embedding simply substitutes $w_{w2vec}.p_t$ to $p_t$ in the LSTM input. 





\paragraph{Semantic decoder}
The semantic decoder is responsible for taking an embedded vector  $s\in \R^S$ and producing a sequence of words in natural language belonging to $\cal P$. It is denoted by $f_s$.
All the semantic decoders we exploit have the structure of an LSTM network. Several semantic decoders will be used for questions, answers and captions in the overall algorithm, but with distinct weights and different inputs. These  LSTMs  have an output predicting word indexes according to a softmax classification network: 
$
p_{t+1} = \mathrm{Softmax}(y_{t})
$
which is re-injected as input of the LSTM at each iteration.

Formally, we can write the semantic decoder as a sequence of words generated by an underlying LSTM first order dynamic process with observations $p_t$ as $y_t=\mathrm{LSTM}_s(y_{t-1}, p_t)$.
At time $t$ the input receives the word generated at the previous state $p_{t}$, and predicts the next word of the sentence $p_{t+1}$. When the word  $w_{0}$ meaning "full stop" is generated at time $T_s$, it ends the generation. The global decoding is therefore a sequence of words of length $T_s - 1$:
$f_s(s;w) = \{p_{t}\}_{t=1:T_{s}-1}$
with the initial state of the LSTM being the embedded vector to decode ($y_0 =s$) and the first input being null ($p_0 = 0$). The symbol $w$ refers to the learned weights of the LSTM and the softmax weights , and is different for each type of textual data that will be generated (captions, questions and answers).



\subsection{Captioning Model\label{sec:cap}}

The visual captioning part of the model is used as an external source of knowledge, and is learned in a separate phase. It takes an image ($I$) and produces a sentence in natural language describing the image. We used an approach inspired by the winner  of the COCO-2015 Image Captioning challenge \cite{Vinyals:2016jd}.  This approach, trainable end-to-end, combines a CNN image embedder with a LSTM-based Sentence Generator (Fig.~\ref{fig:blocks}(b)). 
More formally, it combines an image encoder and a semantic decoder, such are described previously, which can be written as: $c(I) = f_s(f_I(I;w_I) ; w_{cg})$ where $w_{cg}$ is the specific set of learned weights of the decoder.
This caption model acts as an oracle, providing semantic information for the Visual Discriminative Question Selection component, and to the Semantic Representation Encoder.

\subsection{Visual Question Answering Model \label{sec:vqa}}

The VQA model is the second of the two components of our model used as oracles to provide additional information on images.
Its role is to answer independent free-form questions about an image. It receives questions ($Q$) in natural language and an image ($I$) and provides answers in natural language ($a(Q,I)$). Our problem is slightly different from standard VQA because the VQA model now has to answer questions sequentially from a dialog. It means that the question $Q_k$ can be based on the answer of the previous questions $Q_{k-1}$ and answers $a(Q_{k-1},I)$. We first present the formulation of a standard VQA and then show how to extend it so it can answer questions from a dialog.

Inspired by \cite{antol2015vqa}, our  VQA model combines two encoders and one decoder of those described previously (Fig.~\ref{fig:blocks}-a):
\begin{equation}
a(Q,I) = f_s(f_I(I;w_{Iqa})  \odot f_p(Q;w_{pqa}) ; w_{qa})
\label{eq:vqa1}
\end{equation}
The fusion between image and question embeddings is done by an element wise product ($\odot$) between the two embeddings,  as proposed by \cite{antol2015vqa}.



We now consider the case where questions are extracted from a dialog by extending~Eq~(\ref{eq:vqa1}), where $k$ represents the k-th step of the dialog. We introduce another term $h_k$ in the element-wise product to encode the history of the dialog as:
\begin{equation}
a_k(Q_k,I) = f_s(f_I(I;w_{Iqa})  \odot f_p(Q_k;w_{pqa})  \odot h_k  ; w_{qa})
\label{eq:vqa2}
\end{equation}
The history $h_k$ is simply computed as the mean of the previously asked questions/answers, and encoded using $f_p$. This state integrates past questions and is expected to help the answering process. We tried other schemes to summarize history (concatenation, LSTM) without clear performance increase.

We prefer the  VQA model to the Visual Dialog model~\cite{das2016visual}, as this latter is optimized for the task of image guessing, while we want to fine-tune the question/answer sequence for different tasks (multi-label prediction and image retrieval).



\subsection{Discriminative Question Generation}
This part of the model is responsible for taking an image and a caption -- which is considered as the basic semantic representation of the image -- and produces a sequence of questions/answers that are complementary to the caption for a specific task (multi-label classification or retrieval). 

The caption describing the image is generated using the captioning model presented in Section \ref{sec:cap}, and denoted $c(I)\in \cal P$. This caption is encoded with $f_p(c(I);w_{p}) \in \R^S$. Image and caption embeddings are then combined by an element-wise product $f_p(c(I))  \odot f_I(I)$ used as an initial encoded representation of the image.

This representation is then updated iteratively by asking and answering question, one by one, hence  iteratively proposing a list of discriminative questions. Again, we use a  LSTM network (Fig.~\ref{fig:blocks}-c), but instead of providing a word at each iteration as 
for $f_s(s;w)$
we inject a question/answer $[\tilde{q}_k,\tilde{a}_k]$ pair encoded in a vector space from the natural language question/answer $[Q_k,A_k]$ using
$f_p$,
with initial memory $y_0 = f_I(I) \odot f_p(c(I))$ and initial input $\tilde{q}_0 = \tilde{a}_0 = 0$:
$
y_k =\mathrm{LSTM}_q(y_{k-1}, [\tilde{q}_k,\tilde{a}_k])
$
The actual questions are then decoded from the inner LSTM memory $y_k$ and fed to the VQA model to obtain the answer using Eq.(\ref{eq:vqa2}):
$Q_{k+1}=f_s(y_k;w_{sq})$, and, 
$A_{k+1} = a_{k+1}(Q_{k+1},I)$

Using this iterative process, we generate, for each image, a sequence of questions-answers refining the initial caption:
$
f_q(I; w_q) = \{Q_k, A_k\}_{k=1:K}
$
where $w_q$ is the set of weights of the underlying LSTM network of the previous equation, 
and $K$ is an arbitrary number of questions.

\subsection{Semantic Representation Encoder}

Our objective is to evaluate the feasibility of substituting a rich semantic representation to an image and achieve comparable performance than an image feature based approach, for several computer vision tasks. This representation has to be specifically generated to the target task, to be efficient.

Once again, many modern computer vision approaches relying on a learning phase require that data are given as fixed dimension vectors. The role of the module described here is to encode the rich semantic representation in $\R^S$ to feed  the retrieval or the multi-label classification task.

The encoder  makes use of a LSTM network
where the question/answer sequence $\{Q_k, A_k\}_{k=1:K}$ is used as input, $y_k = \mathrm{LSTM}_e(y_{k-1}, [\tilde{q}_k, \tilde{a}_k])$ and the initial memory state $y_0 $ is equal to the encoded caption $f_p(c(I);w_p)$ (Fig.~\ref{fig:blocks}(d)).

If $w_e$ is the set of weights from the underlying LSTM, the rich semantic representation $y_K$ is  encoded as:
$ y_K = f_{enc}\left(\left[\{Q_k,A_k\}_{k=1:K}, c(I)\right];w_e\right)$.





\subsection{Training the model\label{sec:training}}

The global training is divided in two phases. The first phase learns the two so-called oracles independently: image captioning and VQA. The second phase learns to generate the visual quiz, the image encoder and the semantic encoder jointly for a specific task, based on information provided by the two oracles.

The parameters of the VQA, namely  $w_{Iqa}$ and $w_{pqa}$ for the encoders, and $w_{qa}$ (Eq.~\ref{eq:vqa2})  are learned in a supervised way on the Visual Dialog dataset \cite{das2016visual}. The learning criterion consists in comparing the answer words generated by the model with those of the ground truth for each element of the sequence of questions, and is measured by a  cross entropy.

The captioning is also learned by comparing each word sequentially generated by the algorithm to a ground truth caption, and is also measured by a cross entropy loss.
 
The question/answer generator $f_q(I; w_q)$ and the semantic representation encoder $f_{enc}$
are learned jointly end to end. Each of the modules manages its own loss: for the question generator, the sequence of questions is compared to the ground truth of questions associated with each image using a cross entropy at each iteration. The semantic encoding, however, is specifically evaluated by a task-dependent loss: a cross entropy loss for each potential label for the multi-label classification task, a ranking loss for the image retrieval task. When the question generation model converges, only the task-dependent loss is kept in order to fine-tune the question selection part.

The retrieval loss is a bit more complex than the others (cross entropy). Basically, it is based on the assumption that ground truth captions are the most informative image representations and that any other representation should follow the same similarity ranking as captions provide.  We follow the approach proposed in \cite{Gordo_2017_CVPR} to define the retrieval loss as a function of triplet data $q$, $d^+$ (positive pair)  and $d^-$ (negative pair) to be $L(q, d_{+}, d_{-}) = \max(0, m - \phi(q)^{T}\phi(d^+) +   \phi(q)^{T}\phi(d^-)  )$ 
where $q$ and $d^+$ are expected to be more similar than $q$ and $d^-$, and $\phi$ is the representation function to be learned, \ie the output of $f_{enc}$, 
and $m$ is a free coefficient playing the role of a margin. The reference similarity comparison is computed from ground truth captions using \textit{tf-idf} representations, as suggested by \cite{Gordo_2017_CVPR}.

%% file: 6-experiments.tex
\section{Experiments}
\label{sec:experiments}


We validated the proposed method on 2 tasks: i) content based image retrieval (CBIR) based on semantics, where queries are related to the semantic content of the images -- which is more general and harder than searching for visually similar images. We adopted the evaluation protocol proposed by Gordo \etal \cite{Gordo_2017_CVPR}. It uses captions as a proxy for semantic similarity and compares \textit{tf-idf} representations of captions, and measures the performance as the normalized discounted cumulative gain (NDCG), which can be seen as a weighted mean average precision, the relevance of one item with respect to the query being the dot product between their tf-idf representations. 
ii) Multi-label image classification: each image can be assigned to different classes (labels), generally indicating the presence of the object type represented by the label. Per-class average precision and mAP are the performance metrics for this task.

Both series of experiments are done on the Visual Dialog dataset \cite{das2016visual}, relying on images of MS COCO~\cite{lin2014microsoft}.
Each image is annotated with 1 caption and 1 dialog (10 questions and answers), for a total of 1.2M questions-answers.
Ground truth Dialog has been made in order to retrieve a query image from a pool of candidate images. A dialog should visually describe a query image and be suitable for retrieval and classification tasks.
We use the standard train split for learning and validation split for testing, as the test set is not publicly available.

Our approach has several hyper-parameters: the word embedding size, $LSTM$ state size, learning rate, $m$.
They are obtained through cross-validation.
In this procedure, 20\% of training data is considered as validation set, allowing to choose the hyper-parameters maximizing the NDCG/mean average precision on this so-obtained validation set. 
In practice, typical value for $LSTM$ state size (resp. embedding size) is 512 (resp. 200). The margin $m$ is in the range [1.0-2.0].
Model parameters are initialized according to a centered Gaussian distribution ($\sigma= 0.02$). They are optimized with the Adam solver \cite{kingma2014adam} with a cross-validated learning rate (typically of $10^{-4}$), using mini-batches of size 128.
In order to avoid over-fitting, we use dropout  \cite{srivastava2014dropout} for each layer (probability of a drop of 0.2 for the input layers and of 0.5 for the hidden layers). Both oracles (captioning and VQA) are fine-tuned on the tasks. Finally, while it would be interesting to average the performance on several runs, in order to evaluate the stability of the approach, this would be prohibitive in terms of computational time. In practice, we have observed that the performance is very stable and does not depend on  initialization.


\subsection{Experiments on Semantic Image Retrieval}

%
%
\begin{table}[tb]
		\begin{minipage}[t]{0.4\textwidth}

		\caption{NDCG on semantic retrieval. Performance/Area Under the Curve for different values of R.}
		\label{NDCG}
		\resizebox{\textwidth}{!}{\begin{tabular}{l|lll|l}\hline
			Method / R           &  8 &  32 & 128 &  AUC   \\ \hline\hline
			
			$f_i(I)$ + ML (baseline)   &  45.8& 51.7 & 59.3& 69.7 \\ \hline
			$I$ + \cite{Gordo_2017_CVPR}   & 47.6  & 55.9 & 62.3& 72.7 \\ 
			$\{I, c(I)$\} + \cite{Gordo_2017_CVPR}  & 57.0 & 58.5  & 63.3 & 75.1 \\ \hline
			Our approach $f_{enc}(I)$ &  \bf{59.3} & \bf{61.7} & \bf{67.1} & \bf{79.9} \\\hline
		\end{tabular}}
	\end{minipage}
~\begin{minipage}[t]{0.6\textwidth}
	\setlength\tabcolsep{3.5pt}
		\centering
		\caption{Semantic retrieval. NDCG/AUC after removing some components of the model.}
		\label{abla}
		\resizebox{\textwidth}{!}{\begin{tabular}{l|lll|l}\hline
			Modality / R           &  8 &  32 & 128 &  AUC   \\ \hline\hline
			$c(I)$   & 55.1  & 56.3 & 62.4& 73.6 \\ 
			$\{Q_{k}, A_{k}\}_{1:10}$ generic & 41.8 & 50.4  & 57.7 & 65.7 \\
			$\{Q_{k}, A_{k}\}_{1:10}$ task adapted & 45.8 & 55.7  & 60.0 & 71.9 \\
            $tf-idf \{c(I), \{Q_{k}, A_{k}\}_{1:10}\}$ &  54.9 & 57.2 & 63.4 & 75.1 \\\hline
			Our approach $f_{enc}(I)$ &  \bf{59.3} & \bf{61.7} & \bf{67.1} & \bf{79.9} \\\hline
		\end{tabular}}
\end{minipage}
\end{table}


We now evaluate our approach on semantic content-based retrieval, where the images sharing similar semantic content with an image query have to be returned by the system. As described before, the retrieval loss is optimized with  triplets: an image query and two similar/dissimilar images.
For triplet selection, we applied hard negative mining by sampling images according to the loss value (larger loss meaning higher probability to be selected). We found hard negative mining to be useful in our experiments. 

Table \ref{NDCG} reports the NDCG performance for 3 values of R (R=k means that the top k images are considered for computing the NDCG),  and the area under the curve (for~R between 1 and 128) on 4 different models. The visual baseline exploits a similarity metric between image features extracted from the FC7 layer of a VGG19 network, which is learned on the train set using the same triplet approach as described in Section \ref{sec:training}. $I$ + \cite{Gordo_2017_CVPR} corresponds to the visual embedding noted (V, V) in \cite{Gordo_2017_CVPR}. $\{I, c(I)$\} + \cite{Gordo_2017_CVPR} is the joint visual and textual embedding (V+T, V+T) with the difference that we don't feed the ground truth captions but the generated one, for fair comparison.

We observed that the area under the curve improves by +4.8\% with our semantic bottleneck approach  compared to the image feature similarity approach. We stress here that, unlike \cite{Gordo_2017_CVPR}, we only exploit a semantic representation and not image features. 

Empirical results of Table \ref{abla} show the usefulness of our semantic encoder. Indeed, with the same modalities (caption, questions and answers), $tf-idf \{c(I), \{Q_{k}, A_{k}\}_{1:10}\}$ performs 4.8\% lower.  Table \ref{abla} also shows the importance of adapting the VQA oracle to the task with +6.2\% gain compared to a generic oracle not fine-tuned to the task.

\subsection{Experiments on Multi-Label Classification}

With the MS COCO \cite{lin2014microsoft} dataset, each image is labeled with multi-object labels (80 object categories), representing the presence of broad concepts such as animal, vehicle, person, \etc in the image. 
For the  baseline approach, we used image features provided by a VGG-VeryDeep-19 network \cite{simonyan2014very} pre-trained on ImageNet  \cite{russakovsky2015imagenet} with weights kept frozen up to the 4,096-dim top-layer hidden unit activations (fc7), and fed to a final softmax layer learned on the common training set.

\begin{table}[tb]
\caption{Multi-label classification performance.} 
    \centering
 \begin{tabular}{l|l}
Modality & mAP \\ \hline\hline
  $f_i(I)$ (baseline)      & 61.1     \\ \hline
$c(I)$       & 51.6     \\
$\{Q_{k}, A_{k}\}_{1:10}$    & 49.9     \\
$f_{enc}(I)$   & 56.0     \\
$\{I, f_{enc}(I)\}$  & \bf{64.2}    
\end{tabular}

\label{fig:classification}

\end{table}

Table~\ref{fig:classification} (bottom) reports the per-class mean average precision for the visual baseline and the various components of our model.
Our fully semantic approach $f_{enc}(I)$  underperforms only by 5\% the baseline. This is quite encouraging as in our setting the image is only encoded by a caption and 10 questions/answers. 
The main advantage of our model is that one can have access to the intermediate semantic representation for inspection, and may provide an explanation of the good or bad result (see section \ref{sec:bottleneck}). 
Fig. \ref{fig:classification} also reports the performance given by (i) generated caption $c(I)$ only, (ii) questions/answers $\{Q_{k}, A_{k}\}_{1:10}$ only.
These experiments shows that captions are more discriminative than questions/answers (+ 1,7\%), at least given the way they are generated. We also report the performance obtained by combining our image representation  with image features (denoted as $\{I, f_{enc}(I)\}$ ). This configuration gives the best performance (+8,2\%)  and outperforms the  baseline (+3.1\%). 
As a sanity check, we also computed the mAP when using ground truth annotations for both the captions and the VQA. We obtained a performance of 72.2\%, meaning that with good oracles it's possible for our semantic bottleneck to obtain a performance better than with images (61.1\%).

\subsection{Semantic Bottleneck Analysis}
\label{sec:bottleneck}

\begin{figure}[tb]
\centering
\includegraphics[scale=0.30]{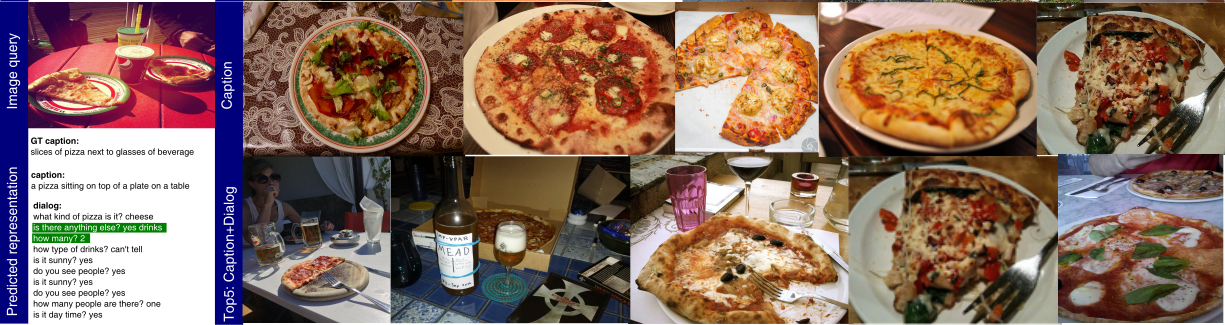}
\caption{Combining captions and dialogs: query (top-left images),  generated captions,  task-specific dialogs,   images retrieved using the caption (first rows) and those given by our model (second rows). Dialogs allowed to detect important complementary
}
\label{fig:captions_and_dialogs}
\end{figure}

This section aims at giving some insights on i) why the performance is improved by combining captions and dialogs and ii) why making the semantic bottleneck adapted to the task improves the performance.

Regarding the first point, we did a qualitative analysis of the outputs of the semantic retrieval task, by  comparing the relevance of the first ranked images when adding the dialogs to the captions (see Figure \ref{fig:captions_and_dialogs}). 
The Figure gives both the caption and the dialog automatically generated, as well as the images ranked first accordingly to the caption (first rows) and accordingly to our model combining the caption and the dialog. We marked in green the  important complementary information added by the dialog. 
The dialog was able to detect drinks as an important feature of the image.

\begin{figure}[tb]
\centering
\includegraphics[scale=0.30]{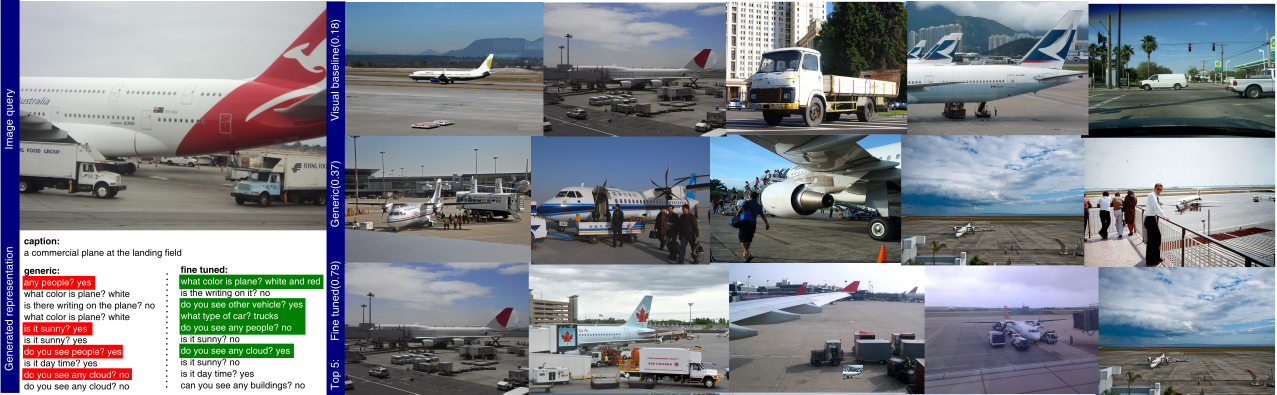}
\caption{Adapting  dialogs to  tasks: query (top-left images),  generated captions, generic and  task-specific dialogs,  images retrieved using the caption and generic dialog (first rows), and those given by our model (second rows).}
\label{fig:fine_tuning_dialogs}
\end{figure}

\begin{figure}[htb]
\centering
\includegraphics[scale=0.24]{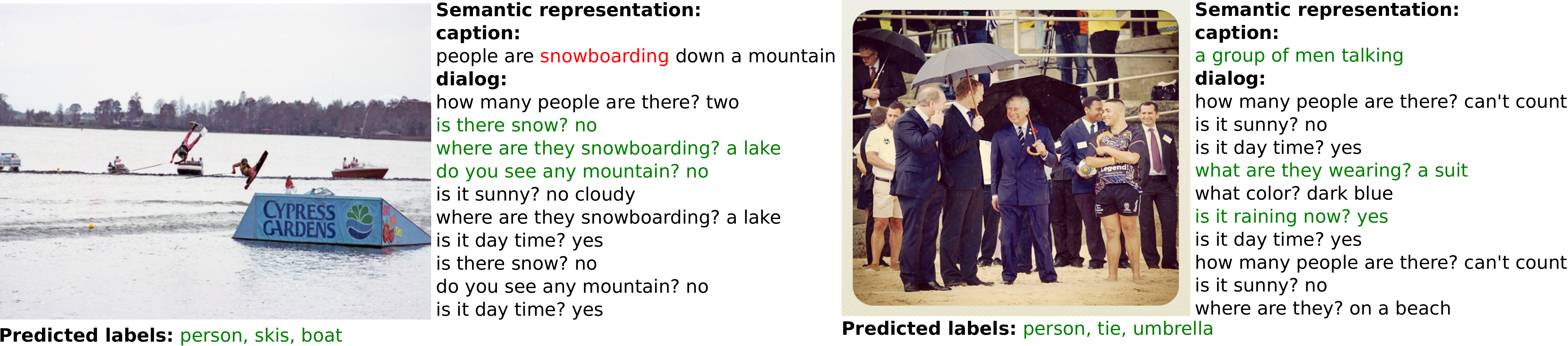}
\caption{Left-hand side: incorrect caption corrected by the dialog. Right-hand side: objects missing from the captions discovered by asking relevant questions.}
\label{fig:fusion}
\end{figure}

Regarding the second point, we compared the quality of the retrieval with and without adapting the dialogs to the task. Figure \ref{fig:fine_tuning_dialogs} illustrates our observations by giving 
 both  captions and dialogs automatically generated, as well as the images ranked first accordingly to captions combined with  generic dialogs (first rows) and with  task adapted dialogs (second rows). We marked in red the questions/answers that we found not relevant to the query in the generic dialog and in green those that have been given by the task adapted dialog to emphasize the complementary information they bring.
The dialog was able to identify vehicles as an important feature of the image.
Fig.~\ref{fig:fusion} illustrates the same type of caption correction by the dialog for the multi-label classification task. 

Generated captions are, in general, brief and consistent with the images (see examples of Fig.~\ref{fig:fine_tuning_dialogs}, \ref{fig:fusion} and \ref{fig:failure_predic}). Because we chose a simple sampling strategy (in order to have a trade-off between computation and interpretation) a few captions are syntactically incorrect. We argue that this should not impact the performance, as the generated captions reflect the image content. We also observed that several questions are repeated. While question repetition is not as critical as it is for the actual dialog generation, it can be overcome if needed by encoding the question history ($h_k$ in Eq.~(6)), for instance by explicitly penalizing repetitions in the LSTM criterion, or, by exploiting a reinforcement learning approach such as in \cite{DasKMLB17}.


Table \ref{tab:stats} illustrates the effect of fine-tuning question generation by showing the percentage of time each word in the first row occurs in a dialog, across the two tasks ('classes' means any of the object class names). We observed that the generated dialogs of the classification task contain more verbs that can be associated to the presence of object classes (eating $\Rightarrow$ food classes, playing~$\Rightarrow$ sport classes, wearing $\Rightarrow$ clothes classes). Generated dialogs for the retrieval task contain more words 
characterizing the scene (in/outdoor, day/night) or referencing specific object features (color, how many).

 \begin{table}[tb]
     \centering
\noindent\resizebox{\textwidth}{!}{
\begin{tabular}{l|l|llll|llll}
task/words   & classes    & playing   & eating   & wearing           & doing              & color           & how many        & in/outdoor & day/night          \\ \hline
classification & \textcolor{white}{\cellcolor{blue!88} 88\%} & \textcolor{black}{\cellcolor{blue!19} 19\% }  & \textcolor{black}{\cellcolor{blue!39} 39\% }& \textcolor{black}{\cellcolor{blue!29} 29\% }  & \textcolor{black}{\cellcolor{red!14} 14\% } & \textcolor{white}{\cellcolor{red!42} 42\% }  & \textcolor{white}{\cellcolor{red!39} 39\% }  & \textcolor{white}{\cellcolor{red!42} 42\% } & \textcolor{white}{\cellcolor{red!53} 53\% } \\
retrieval  & \textcolor{white}{\cellcolor{blue!78} 78\%} & \textcolor{black}{\cellcolor{blue!14} 14\%} & \textcolor{black}{\cellcolor{blue!28} 28\% }& \textcolor{black}{\cellcolor{blue!21} 21\%}    & \textcolor{black}{\cellcolor{red!16} 16\%} & \textcolor{white}{\cellcolor{red!85} 85\%} & \textcolor{white}{\cellcolor{red!81} 81\%} & \textcolor{white}{\cellcolor{red!54} 54\% }& \textcolor{white}{\cellcolor{red!79} 79\%}
\end{tabular}
}\caption{Statistics of generated words after fine tuning the generators to the tasks\label{tab:stats}}
\end{table}

\begin{figure}[tb]
\centering
\includegraphics[scale=0.2]{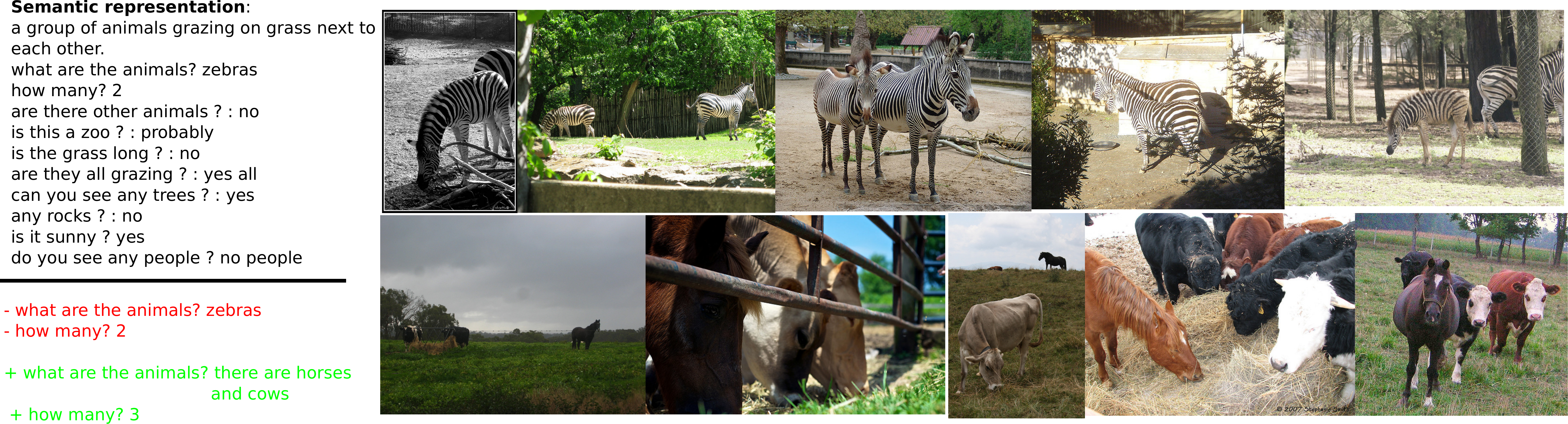}
\caption{Incrementally updating the representation.}
\label{updating_representation}
\end{figure}

We also made experiments showing how the semantic bottleneck can be modified manually to make image search more interactive. Fig.~\ref{updating_representation} shows an example where we changed 2 oracle answers (zebras becomes cows and their number is increased by one). The 2nd row depicts the impact of this modification.

\subsection{Evaluating Failure Predictions}

\begin{figure}[tb]
\centering
\includegraphics[scale=0.24]{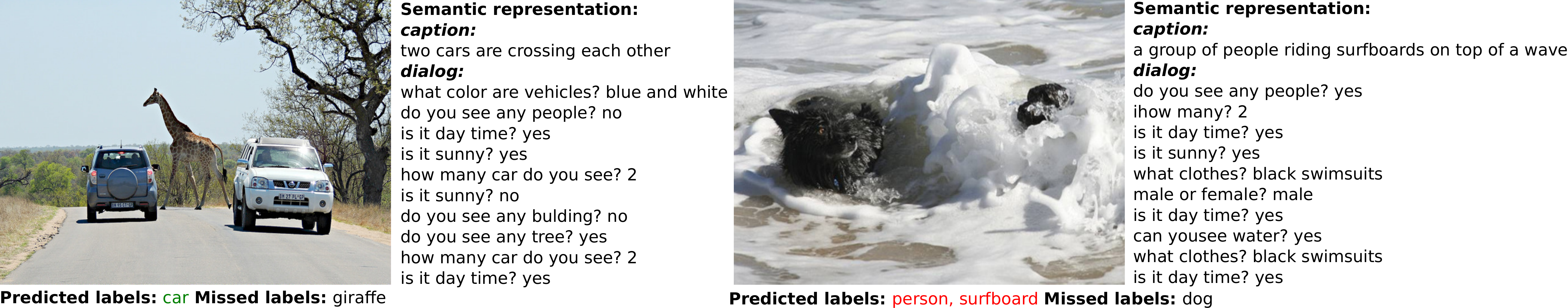}
\caption{Predicting failure cases from the proposed semantic representation.  {Left-hand side:} caption and Q/A are consistent but not rich enough to predict the 'giraffe' label.  {Right-hand side:} the semantic representation is incorrect leading to the inference of  erroneous labels. In such cases, the bottleneck representation can be used for debugging. 
}
\label{fig:failure_predic}
\end{figure}

The potential capacity of the semantic bottleneck to detect failure in the prediction process is illustrated by Fig.~\ref{fig:failure_predic}. Failure is detected when the representation contains incorrect semantic information --- the caption or dialog are wrong --- or insufficient information for further inference.
We focus our evaluation on multi-label image classification, since a clear definition of failure in the case of content based image retrieval is complex, can be subjective (how decide if images are completely dissimilar from the request?) and task oriented (what are retrieved images used for?). 
We developed two evaluation protocols: one with humans in the loop, judging the semantic bottleneck capacity to predict consistent labels, and an automatic model optimized to predict success or failure for each class. We compare our approaches to a baseline based on score prediction thresholding.
\begin{table}[tb]
		\begin{minipage}[t]{0.5\textwidth}
				\centering \caption{Failure prediction statistics. }
				\begin{tabular}{r|cc|cc|}
					& \multicolumn{2}{c|}{\em false negative} & \multicolumn{2}{c}{\em false positive} \\ \hline
					& \#true & \#predicted & \#true & \#predicted\\ \hline
					GT.    & 614  & - & 588 & -   \\ \hline 
					users   & 308 & 379 &  213  & 485\\ \hline 
						classifier   & 250 & 490  &   180 &  530\\ \hline
				\end{tabular}
                \label{table:stats_result}
                \end{minipage}
		\begin{minipage}[t]{0.5\textwidth}
				\centering \caption{Multi-label classification.
				}
				\begin{tabular}{r|cc|cc|}
					& \multicolumn{2}{c|}{\em label} 	& \multicolumn{2}{c}{\em image}  \\ \hline
					& mAP   & \% & mAP   & \% \\ \hline
					no selection    & 54.3 & 100  & 54.3 & 100 \\ \hline
					users    & 84.2&  96  & 86.1 & 53\\  \hline 
					classifier    & 79.8 & 93 & 81.7 & 49\\   \hline     				   
                    conf. thresh.  & 66.1 & 93 & 73.5 & 49\\  \hline 
				\end{tabular}
                 \label{table:avg_result}
                 \end{minipage}
\end{table}
In order to evaluate the semantic bottleneck capacity, we first train our model for a multi-label classification task and extract the generated semantic representation (caption and dialog) and class prediction.

\noindent\emph{Human based failure prediction study.}
For 1000 randomly chosen test images,  users were instructed to evaluate the capacity of the semantic bottleneck to contain enough information to predict the correct classes. The image and the generated semantic representation are shown to the users, which can select for each of the 80 labels of MS-COCO 1 among 3 cases: i) \textit{false negative}. The semantic representation missed the label (\eg caption and dialog do not mention about the horse in the picture). ii) \textit{false positive}. The semantic representation hallucinates the object (\eg seeing a car in a kitchen scene). iii) \textit{correct}. The algorithm has succeeded to predict the label, either its absence or its presence. Table \ref{table:stats_result} shows failure cases of the multi-label classification (614 false negative and 588 false positive). Human subjects were able to identify half of the failures (308/614 FN and 213/588 FP) with a precision of $\approx 60\%$ (308/379 and 213/485).

Failure detection can also be evaluated through two other sets of experiments: \textit{Label rejection}: suspicious labels are rejected, others are kept. \textit{Image rejection}: when there is a suspicious label, the image is rejected. Table~\ref{table:avg_result} shows both experiments, and reads as follows: classification performance is of 54.3\% when evaluating on 100\% of the test set. When user rejects 4\% of the labels, the performance goes to 84.2\%. When our rejection algorithm keeps 93\% of the data, the performance improves to 79.8\%, which is close to human performance. We see a strong improvement for both our methods. Failure prediction improves the average precision of 30\% percent with 4\% of deleted image in average for each class. 

We also proposed two automatic algorithms for failure prediction. The first one, referenced as 'classifier' in Tables~\ref{table:stats_result} and~\ref{table:avg_result}, is based on an independent ternary linear classifier for each class with 3 possible outputs: $\mathtt{correct}$, $\mathtt{FN}$, $\mathtt{FP}$. The input is the image $I$ concatenated with the last hidden state of the semantic representation encoder $[\{Q_k,A_k\}_{k=1:K}, c(I)]$. The ground truth is built by comparing the output from the multi-label classification and the true classes. The model is optimized using a cross entropy loss.
It is less accurate (can detect $\approx 41\%$ of false positive with $\approx 51\%$ of precision) but has the advantage of reducing human effort.
We also show in the last row of Table~\ref{table:avg_result} the performance of a second algorithm consistng in thresholding the confidence score outputed by the multi-label classifier for each label, and tuned to reach the same rejection rate as the other failure detection algorithm. This confidence thresholding algorithm gives a smaller performance increase after rejection.


%% file: 7-conclusions.tex
\section{Conclusions}
In this paper we have introduced a novel method for representing images with semantic information expressed in natural language. Our primary motivation was to question the possibility of introducing an intelligible bottleneck in the processing pipeline. We showed that by combining and adapting several state-of-the-art techniques, our approach is able to generate rich textual descriptions that can be substituted for images in two vision tasks: semantic content based image retrieval, and multi-label classification. We quantitatively evaluated the usage of this semantic bottleneck as a diagnosis tool to detect failure in the prediction process, which we think contributes to a clearer metric of explainability, a key concern to mature artificial intelligence.  

